\documentclass[11pt]{article}

\usepackage[preprint]{acl}

\usepackage{times}
\usepackage{latexsym}

\usepackage[T1]{fontenc}

\usepackage[utf8]{inputenc}

\usepackage{microtype}

\usepackage{inconsolata}

\usepackage{graphicx}

\usepackage[ruled,vlined]{algorithm2e}
\usepackage{subcaption}
\usepackage{booktabs}
\usepackage{colortbl}
\usepackage{multirow}
\usepackage{bm}
\usepackage[most]{tcolorbox}

\newcommand{\myrq}[2]{%
    \begin{tcolorbox}[colback=blue!10, colframe=cyan!25!black, boxrule=0.5pt]
        \textbf{RQ#1:} #2
    \end{tcolorbox}
}

\newcommand{\pv}{\textit{p}-value}
\newcommand{\pvs}{\textit{p}-values}

\newcommand{\mae}{$\Gamma_{\rm MAE}$}

\newcommand{\hide}[1]{}

\title{Thesis Proposal: Toward a Human-Centered and Perspective-Aware Framework for Reproducible ML Evaluation and AI Alignment}

\author{Deepak Pandita \\
  Rochester Institute of Technology \\
  \texttt{deepak@mail.rit.edu} \\\And
  Christopher M. Homan \\
  Rochester Institute of Technology \\
  \texttt{cmh@cs.rit.edu} \\}

\begin{document}
\maketitle
\begin{abstract}
Humans play a vital role at every stage of AI development, from data collection and curation to model development and evaluation. However, humans often disagree with each other and sometimes with themselves over time. It is essential to take disagreement into account when building human-centered AI systems, especially in domains where it is prevalent, such as AI safety, content moderation, or sentiment analysis. Disagreement often arises from subjective human opinion and can vary with one's identity, beliefs, and social environment. Despite this, current LLM evaluation approaches frequently rely on aggregating labels (often via plurality voting) to represent consensus, thereby obscuring minority perspectives. By failing to account for human disagreement, these evaluation methods contribute to the reproducibility crisis in AI. Human feedback is also crucial for ensuring that AI systems align with human values. For these systems to be trustworthy, it is critical to ensure that they reflect diverse human values and perspectives. In this thesis proposal, we present a human-centered and perspective-aware framework for reproducible ML evaluation and AI alignment.
\end{abstract}

\section{Introduction}
\label{sec:introduction}

With the increasing prevalence of AI in modern society, it is of utmost importance that AI systems are reliable and trustworthy. Besides the models being helpful and harmless, a desired property is that the model evaluations are thorough and reproducible. Yet AI remains in a crisis~\citep{baker_1500_2016, Gundersen_Kjensmo_2018, hutson_2018, mieskes-etal-2019-community, Gundersen_2020} in which researchers are not able to reproduce the results of previous studies~\citep{NEURIPS2019_c429429b}. Many factors caused and sustain this crisis: documentation is incomplete; methods, algorithms, and implementations vary; and gold standard data can be unreliable.

We focus here on an overlooked source of unreliability: failing to account for human disagreement and other sources of randomness in ML evaluation. Conventional evaluations treat disagreement, if at all, as nothing more than noise and may aggregate 3--5 labels per item---a number that comes from literature on machine learning~\cite{snow2008cheap}, not machine learning evaluation---via plurality voting to represent consensus, overlooking disagreement~\cite{barile2021toward, davani-etal-2022-dealing}, which is endemic in human responses.

Recently, there has been an increased emphasis on the idea of \textit{perspectivism}\footnote{\url{https://pdai.info/}}, which advocates integrating diverse perspectives in machine learning~\cite{Cabitza_Campagner_Basile_2023}. Perspectivist approaches urge using and publishing disaggregated labels to account for human label variation~\cite{basile-etal-2021-need, prabhakaran-etal-2021-releasing, uma2021learning, plank-2022-problem}.
Consequently, model evaluations must be \textit{perspective-aware} to ensure both trustworthiness and reproducibility.

\paragraph{Thesis Statement} \textit{Accounting for diverse perspectives is critical to improving reproducibility in machine learning evaluation and ensuring pluralistic alignment of large language models.}

Toward this end, we pose the following research questions:

\begin{itemize}
    \item \textbf{RQ1:} \textit{Is it valuable to keep disaggregated responses for each item while comparing two ML models?}
    \item \textbf{RQ2:} \textit{How do we optimize the allocation of a fixed human annotation budget for reproducible evaluation?}
    \item \textbf{RQ3:} \textit{What is the impact of diverse raters on the amount of data needed for reproducibility?}
\end{itemize}

In the case of LLMs, humans continue to play a vital role in all stages of their development. Human feedback is used to align LLMs to reflect human values and preferences via reinforcement learning from human feedback (RLHF)~\cite{christiano_deep_2017, ziegler_fine-tuning_2019, ouyang_training_2022, bai_constitutional_2022}. However, recent research has shown that such alignment can favor specific political ideologies~\cite{santurkar_whose_2023,ceron-etal-2024-beyond,fulay-etal-2024-relationship}. AI systems must represent the values and preferences of diverse groups, not just one of them. Therefore, there is an increased interest in building human-centered AI systems that reflect pluralistic views/values~\cite{gordon2022jury,sorensen_value_2024,sorensen_roadmap_2024,stammbach-etal-2024-aligning}. This remains a key challenge since humans may have diverging viewpoints~\cite{casper_open_2023}. In light of these challenges, we seek to answer the following research questions:

\begin{itemize}
    \item \textbf{RQ4:} \textit{Are some demographic groups more cohesive than others when disclosing their own perceptions of offense and vicarious offense?}
    \item \textbf{RQ5:} \textit{Can textual feedback enhance the performance of LLMs during inference?}
    \item \textbf{RQ6:} \textit{How can we effectively adapt LLMs to represent pluralistic perspectives?}
\end{itemize}

The rest of this manuscript is structured as follows: Section \ref{sec:background} provides necessary background; Sections \ref{sec:reproducible_evaluation} and \ref{sec:alignment} detail our progress in perspective-aware reproducible ML evaluation and human-centered AI alignment, respectively. Finally, Section \ref{sec:conclusion} summarizes the proposal’s contributions and concludes the work.

\section{Background}
\label{sec:background}

\subsection{Human Disagreement}

Subjectivity often leads to disagreement and causes variance in human responses~\cite{basile-etal-2021-need, prabhakaran-etal-2021-releasing, uma2021learning, plank-2022-problem, Cabitza_Campagner_Basile_2023, weerasooriya-etal-2023-vicarious}. Disagreement is also linked to rater identity (race, gender, age, education, and first language) and their beliefs (social, religious, spiritual, and political leaning)~\citep{sap-etal-2019-risk, al-kuwatly-etal-2020-identifying, larimore-etal-2021-reconsidering, sap-etal-2022-annotators, goyal2022your, pei-jurgens-2023-annotator, weerasooriya-etal-2023-vicarious, homan-etal-2024-intersectionality, prabhakaran-etal-2024-grasp}. Typically, responses are aggregated via majority voting to represent consensus, whereas recent work has shown the inadequacy of majority voting for incorporating response variance~\cite{barile2021toward, davani-etal-2022-dealing}.

\subsection{Reproducible ML Evaluation}

Several studies have called attention to the reproducibility crisis in AI and NLP~\cite{Gundersen_Kjensmo_2018, hutson_2018, mieskes-etal-2019-community, Gundersen_2020}. Given the non-deterministic nature of machine learning methods, algorithms, and implementations, even if code is shared, multiple identical training runs of the same deep learning model can produce different models, with different test results~\cite{pham_problems_2020}. \citet{pham_problems_2020} also presented a survey of 901 participants, where $84\%$ were unaware or unsure about the variance caused by different implementations. \citet{arvan-etal-2022-reproducibility-code} conducted a reproducibility study of eight papers published in EMNLP 2021 and achieved a $25\%$ success rate. Therefore, it is vital to account for variance in evaluations.

Human evaluation studies also show a low degree of reproducibility~\cite{belz_non-repeatable_2023}.
The field also faces a pervasive issue of inadequate statistical analysis; statistical significance is often misapplied, and reported outcomes are frequently unreliable~\cite{sogaard-etal-2014-whats, dror-etal-2018-hitchhikers, van-der-lee-etal-2019-best}.

\subsection{Prompt Optimization}

Automatic prompt optimization methods, such as AutoPrompt~\cite{shin-etal-2020-autoprompt} and RLPrompt~\cite{deng-etal-2022-rlprompt}, employ gradient-based search and reinforcement learning techniques, respectively. Other approaches leverage LLMs themselves for prompt generation~\cite{mehta2024promptly, pryzant-etal-2023-automatic, yang2024largelanguagemodelsoptimizers, yang-etal-2022-re3, zhou2022large}. Recent works like Promptomatix~\cite{murthy2025promptomatixautomaticpromptoptimization} and EvoAgentX~\cite{wang2025evoagentxautomatedframeworkevolving} extend this direction by enabling automatic prompt refinement across multiple tasks, workflows, and tools.

\section{Perspective-Aware Reproducible ML Evaluation}
\label{sec:reproducible_evaluation}

\subsection{Related Work in Reproducible ML Evaluation}
\citet{wein-etal-2023-follow} proposed the Variance Estimation Toolkit (VET)\footnote{\url{https://github.com/google-research/vet}} for estimating p-values for comparisons between the results of two systems to determine which is ranked higher. The framework uses null hypothesis significance tests (NHSTs) to demonstrate model improvement while accounting for sampling variance across items and responses per item. It explores which sampling, aggregation, and measurement methods yield the best p-value estimate from a single test set relative to the true/ground-truth p-value.

VET simulator samples responses from a large pool of human raters ($G$) and two machine learning models ($A$ and $B$). To simulate responses for $G$, it uses a \textit{multistage sampling} approach to generate responses for $N$ items with $K$ responses per item. First, for each item, a mean and a standard deviation are sampled from specific uniform distributions. Then, $K$ responses are sampled from a normal distribution parameterized with the sampled mean and standard deviation. The responses for items in models $A$ and $B$ are generated using the same parameters as $G$, but with means perturbed by a small amount $\epsilon$ (chosen uniformly at random over a specific interval) for model $B$. This makes model $A$ a perfect representation of $G$. The data for the null hypothesis is generated by combining the responses for model $A$ and model $B$ into a single set and sampling from this set. NHSTs are then used to estimate the p-values under different metrics and sampling methods.

\citet{homanmany} utilized the VET simulator to study the trade-off between the number of items and responses per item using simulations tailored for foundation models.
The VET simulator's scope is limited to NHST and regression. Furthermore, \citet{homanmany} evaluate the system using just simulations instead of real-world datasets.

\subsection{Preliminary Work in Reproducible ML Evaluation}
\label{sec:vet}

\myrq{1}{Is it valuable to keep disaggregated responses for each item while comparing two ML models?}

Leveraging the VET simulator, we present a human-in-the-loop method~\cite{homan-etal-2026-many}
to estimate the number of items ($N$) and the number of responses per item ($K$) needed for reliable comparison of two ML models under a performance difference of at least $\epsilon$ according to a metric $\Gamma$. We achieve this by computing \pvs\ for existing experimental data comparing the performance of two models against gold data under different experimental conditions. We also extend the simulator to estimate the type-II error rate, allowing for statistical power.

\paragraph{Methods} Given an evaluation dataset $G$, arbitrary $N$ and $K$, $\epsilon > 0$ and metric $\Gamma$ the process has the following steps.
\begin{enumerate}
    \item Fit a two-stage probabilistic \emph{response model} model to $G$.
    \item Use that model via \emph{simulation} to determine p-values for $N$, $K$, $\epsilon$, and $\Gamma$.
\end{enumerate}

First, we fit a dataset to a response model by treating it as a regression task and following the same process as mentioned in \citet{wein-etal-2023-follow}. Then, we use the simulator to generate data for $G$ using the same fitted distribution. Next, we generate the data for model $A$ using the same distribution as $G$, making $A$ an ideal representation of $G$. The data for model $B$ is generated by adding a perturbation $\epsilon$ to $G$. It ensures that model $A$ always outperforms model $B$ with respect to $G$ as measured by any metric $\Gamma$. The \pvs\ should converge to zero as $N$, $K$, and/or $\epsilon$ increase. This process is repeated a large number of times to estimate the \pvs.

\paragraph{Experiments} We utilize seven datasets to conduct our experiments -- MultiDomain Agreement~\cite{leonardelli2021agreeing}, Stanford Toxicity~\cite{kumar2021designing}, Amazon reviews~\cite{zhang2015character}, HS-Brexit~\cite{akhtar2021whose}, ConvAbuse~\cite{cercas-curry-etal-2021-convabuse}, ArMIS~\cite{almanea-poesio-2022-armis}, and Measuring Hate Speech (MHS)~\cite{sachdeva-etal-2022-measuring} all of which contain multiple human annotations per item. 

Our evaluation relies on the following metrics:
\begin{itemize}
\item \textit{Mean absolute error difference} (MAE).
The errors from the per-item mean gold response to the model response averaged over the items.

\item \textit{Item-wise wins} (Wins).
The fraction of items in the test set for which the absolute error of A is smaller than that of B.

\item \textit{Mean EMD difference} (MEMD).
The Earth mover's distance for each item between the system and the gold standard responses, and then take the mean of those item-wise EMDs.
\end{itemize}

\begin{figure}[htb]
\centering 
\begin{subfigure}[]{\linewidth}
\centering
\includegraphics[width=\linewidth]{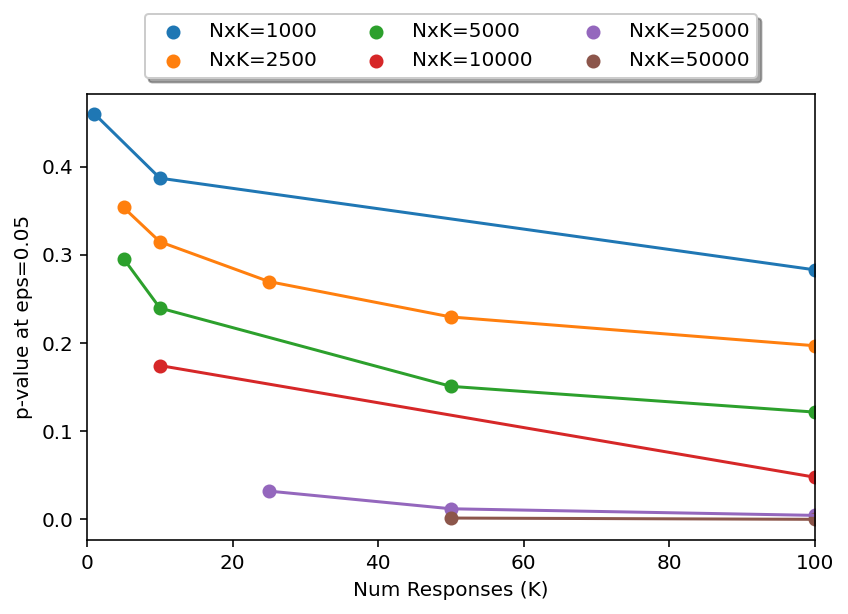}
\caption{Toxicity ($\epsilon=0.05$)}
\label{fig:toxicity_pval_for_NxK}
\end{subfigure}
\hspace{.3cm}
\begin{subfigure}[]{\linewidth}
\centering
\includegraphics[width=\linewidth]{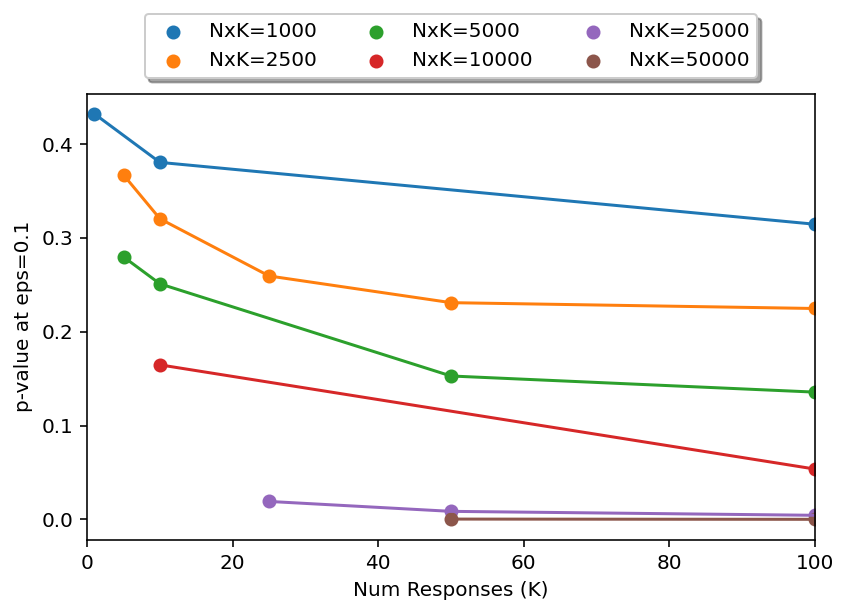}
\caption{MultiDomain ($\epsilon=0.1$)}
\label{fig:mdagreement_pval_for_NxK}
\end{subfigure}
\caption{
\pv\ vs $K$ with \mae\ at various $N \times K$. Each data point is estimated from $10,000$ samples. 
}
\label{fig:pval_for_NxK}
\end{figure}

\paragraph{Results} Our results (Figure~\ref{fig:pval_for_NxK}) demonstrate that trading off items for responses is beneficial at a wide range of ($N\times K$) values, with \pv\ decreasing as $K$ increases. Here \mae\ was used with distortion $\epsilon=0.05$ for Toxicity and $\epsilon=0.1$ for MultiDomain, but similar trends were observed using other metrics, amounts of distortion, as well as different datasets.

Figure~\ref{fig:various_metrics} graphs \pv\ as a function of the number of responses at $\epsilon=0.1$, where the number of items varies such that $N \times K = 2500$, and demonstrates a similar trend across five different metrics. Refer to Appendix \ref{sec:power_analysis} for the results of the power analysis.

The results also suggest that current evaluation practices are not sufficient to confidently assess two models' performance against gold judgments, as using 25,000-50,000 annotations in a test set is rarely seen. Even when using 1000 items, at least 25 raters are needed for models to achieve significance with MAE.

\myrq{2}{How do we optimize the allocation of a fixed human annotation budget ($N\times K$) for reproducible evaluation?}

\begin{figure}[ht]
    \centering
    \includegraphics[width=\columnwidth]{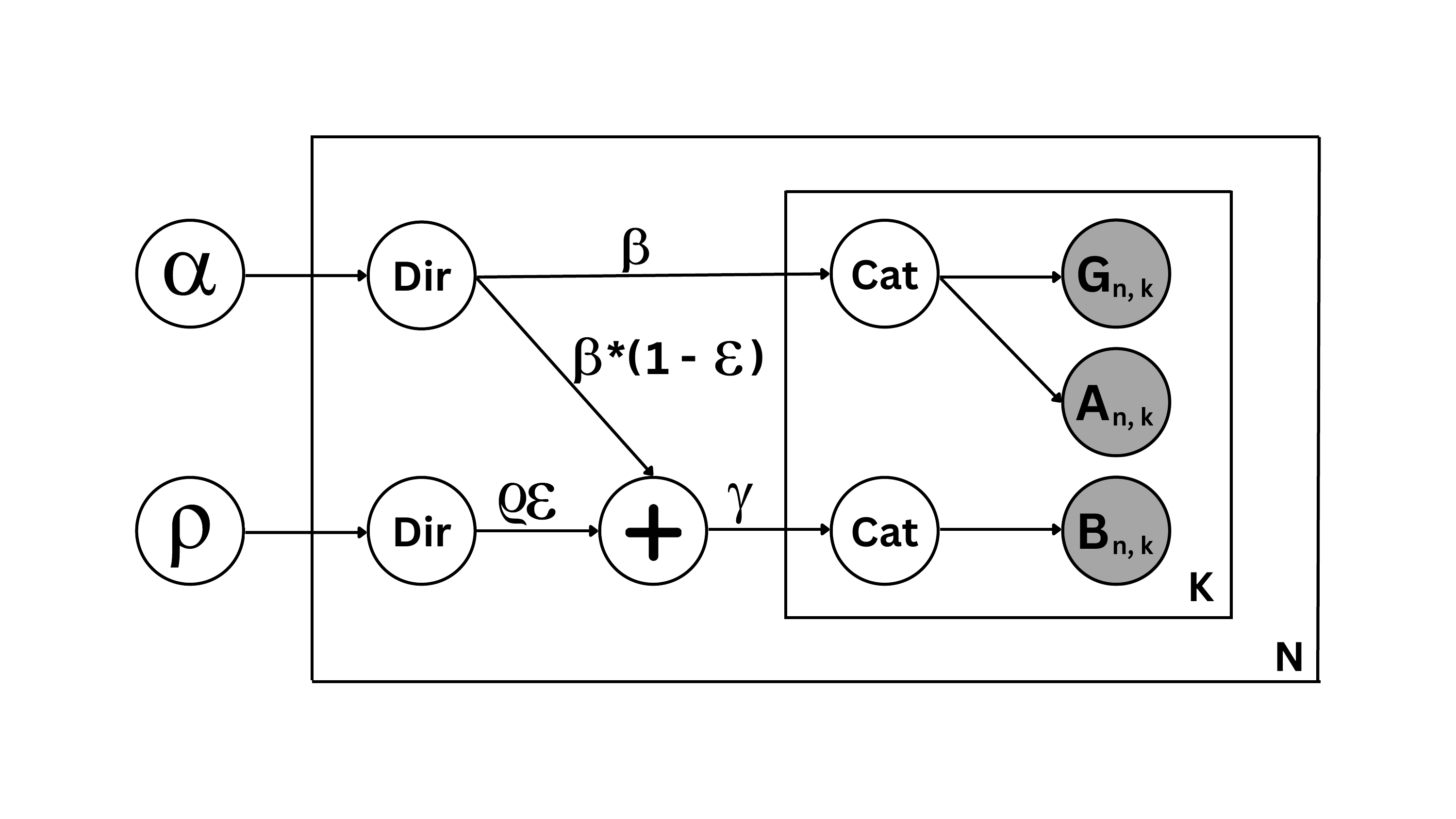}
    \caption{Plate notation for the simulator. Categorical parameters ($\beta$) and noise parameters ($\varrho$) are sampled from two Dirichlet distributions parameterized by $\alpha$ and $\rho$, respectively. Then, responses for $G$ and $A$ are produced by sampling from a categorical distribution parameterized by $\beta$. Responses for $B$ are produced by sampling from a categorical distribution parameterized by $\gamma$, where $\gamma$ is a convex combination of $\beta$ and $\varrho$ controlled by the perturbation parameter $\epsilon$.}
    \label{fig:fig_plate_notation}
\end{figure}

Next, we extend the VET simulator to model categorical data and confidence intervals~\cite{Pandita_Korn_Welty_Homan_2026}.
We adopt a Bayesian approach to model existing datasets and examine the optimization problem of allocating a human annotation budget to a sample of $N$ items, where each item is annotated by $K$ raters, such that the total budget $N \times K$ is fixed. This enables a more robust way of modeling when the sample size is small and allows for maximum \emph{a posteriori} (MAP) fitting of data, versus maximum likelihood estimation (MLE)-based frequentist approaches, which provides regularization.

\paragraph{Methods} The simulator (Figure \ref{fig:fig_plate_notation}) works similarly to earlier work~\cite{wein-etal-2023-follow} except that the responses produced are nominal rather than continuous. The simulator produces gold responses $G$ and the responses for model $A$ by sampling from the same Dirichlet-categorical distribution. For model $B$, the responses are sampled after perturbing the parameters by a small amount $\epsilon$.

\paragraph{Experiments} We use five datasets -- Toxicity~\cite{kumar2021designing}, DICES 350~\cite{NEURIPS2023_a74b697b}, D3code~\cite{davani_d3code_2024}, and Jobs Q1/Q3~\cite{liu-etal-2016-understanding}, for which multiple annotations per item exist, and fit these datasets using MAP. We run experiments for hypothesis testing with different budgets ($N\times K$ = \{100, 250, 500, 1000, 2500, 5000, 10000, 25000, 50000\}) while ranging $K$ from 1 to 500 (in increments of 1 till 10, then 20, then in increments of 20 from 20 onwards)
for different metrics, and $\epsilon$ = \{0.1, 0.2, 0.3, 0.4\}. We use four metrics with four $\epsilon$, yielding 16 sets of 282 experiments for each dataset.
We use the following metrics for our experimentation:

\begin{itemize}
    \item \textit{Accuracy}. Accuracy is the most commonly used metric to compare models against each other. First, take the plurality vote for all items in $A$, $B$, and $G$. Then compute the accuracy for $A$ and $B$ by comparing against $G$.
    
    \item \textit{Total variation} (TV). TV is related to Manhattan or L1 distance. It goes beyond the plurality vote and helps compare probability distributions for soft label evaluation. Compute the frequency of responses for all items in $A$, $B$, and $G$, normalize, and compute the mean Manhattan distance across all items in $A$ and $B$ against $G$.
    
    \item \textit{Wins}. Wins is a meta-metric used for item-level comparison. We use TV as the base metric for Wins, but any other metric can be used. Calculate TV for all items in $A$ and $B$ against $G$, then count the wins of $A$ and $B$, i.e., the number of times $A$ has less TV than $B$ and vice-versa.
    
    \item \textit{KL-Divergence} (KL-Div). KL-Divergence is another frequently used metric for comparing probability distributions. Calculate the frequency of responses for all items in $A$, $B$, and $G$. Then, compute the mean KL-divergence across all items in $A$ and $B$ against $G$.
    
\end{itemize}

\begin{table}[h]
\centering
\scriptsize
\begin{tabular}{l|c|cccc}
 &  & Accuracy & TV & Wins & KL-Div \\
\midrule
 & NK & 2500 & 1000 & 2500 & 1000 \\
Toxicity & \pv\ & 0.012 & 0.015 & 0.012 & 0.022 \\
(M=2) & K & 1 & 120 & 1 & 200 \\
 & $\Delta$ & 0.040 & 0.074 & 0.040 & 0.044 \\
\hline
& NK & 1000 & 500 & 1000 & 1000 \\
DICES & \pv\ & 0.036 & 0.017 & 0.028 & 0.020 \\
(M=3) & K & 1 & 80 & 20 & 300 \\
 & $\Delta$ & 0.055 & 0.063 & 0.346 & 0.082 \\
\hline
& NK & 2500 & 1000 & 2500 & 1000 \\
D3code & \pv\ & 0.037 & 0.020 & 0.024 & 0.022 \\
(M=2) & K & 2 & 140 & 60 & 100 \\
 & $\Delta$ & 0.034 & 0.072 & 0.413 & 0.036 \\
\hline
& NK & 250 & 250 & 250 & 250 \\
JobsQ1 & \pv\ & 0.035 & 0.015 & 0.036 & 0.035 \\
(M=5) & K & 1 & 40 & 1 & 1 \\
 & $\Delta$ & 0.104 & 0.050 & 0.104 & 2.864 \\
\hline
& NK & 500 & 250 & 500 & 500 \\
JobsQ3 & \pv\ & 0.047 & 0.014 & 0.038 & 0.030 \\
(M=12) & K & 100 & 240 & 80 & 500 \\
 & $\Delta$ & 0.595 & 0.024 & 0.868 & 0.182 \\
\end{tabular}
\caption{Minimum \pv, $K$, and corresponding effect size ($\Delta$) for lowest $NK$ with $p<0.05$ ($ \epsilon=0.3$).}
\label{tab:low_k_for_p_lte_05_nk_min_p}
\end{table}

\paragraph{Results} Table \ref{tab:low_k_for_p_lte_05_nk_min_p} shows the results for minimum \pv, $K$, and corresponding effect size ($\Delta$) for lowest $NK$ with $p<0.05$ ($ \epsilon=0.3$). $M$ represents the number of categories in the dataset. Our results suggest that whether or not a tradeoff exists, and where it is, depends much more on the metric used than the data source, and that the metrics behave very differently. They show that the TV metric requires the smallest number of $N \times K$ overall, and that this comes with a small number of $K > 10$.

Our findings demonstrate that increasing $K$ is often a more effective strategy for achieving reliable evaluation than increasing $N$. We discovered, across a diverse set of datasets, that accounting for the full human response distribution can be achieved with a surprisingly modest budget ($N \times K$) of 1000 or less, with $K>10$. Metrics that are more sensitive to the distributional nature of human responses benefit greatly from higher values of $K$.

\subsection{Proposed Work in Reproducible ML Evaluation}

\myrq{3}{What is the impact of diverse raters on the amount of data needed for reproducibility?}

\begin{figure}
\centering
\includegraphics[width=\linewidth]{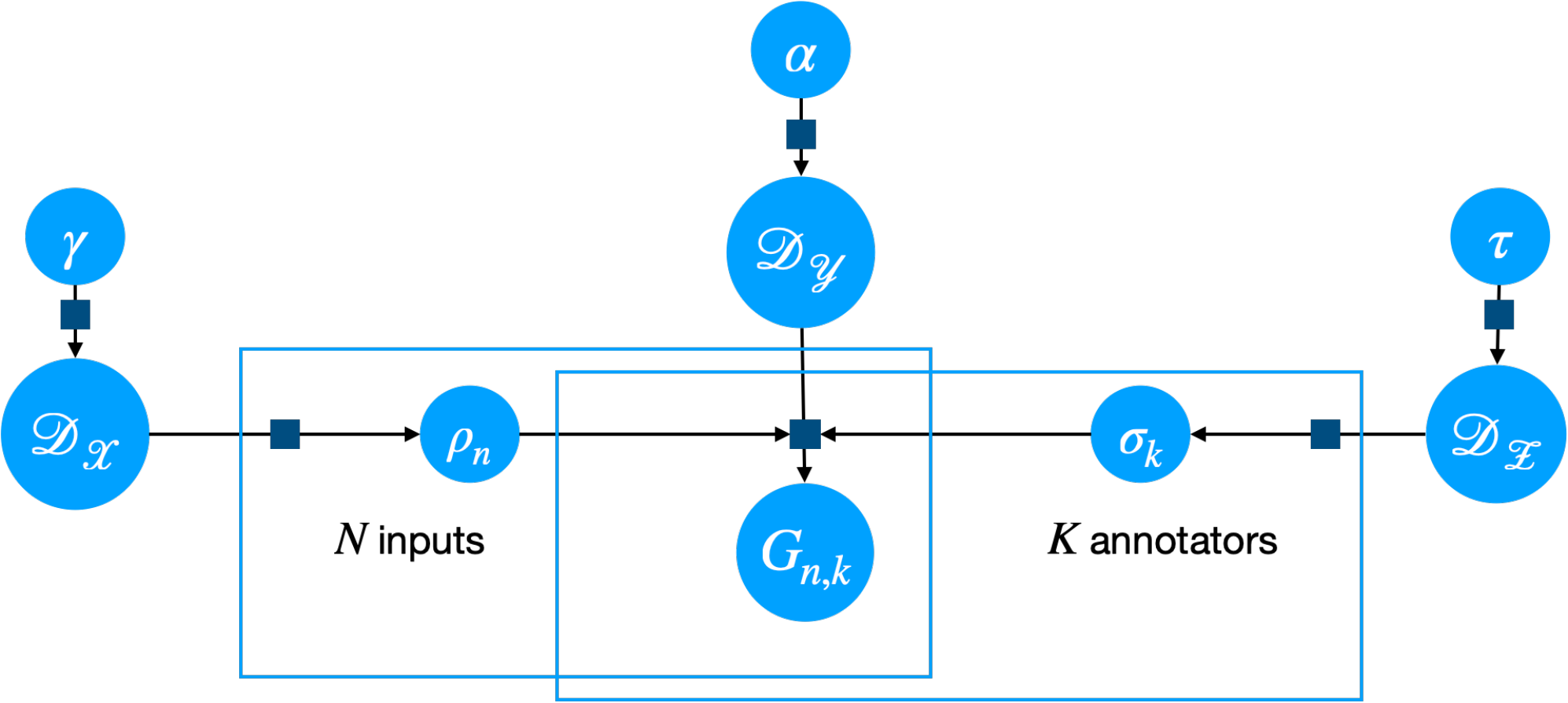}
\caption{Plate diagram for hierarchically modeling individual annotators and input items, simplified from~\cite{weerasooriya-etal-2022-improving}. The approach is similar to Algorithm \ref{alg:alt}, except that in addition to sampling model parameters $\rho_n$ for each item $n$, we also sample parameters $\sigma_k$ for each annotator $k$. Then for each pair $(n,k)$, one response is sampled from $\mathcal{D_Y}(\rho_n, \sigma_k)$, which is now parameterized by \emph{both} $\rho_n$ and $\sigma_k$.}
\label{fig:graphical}
\end{figure}

Currently, the VET simulator assumes that the responses for one input item are independent of those of any other item, given the inputs. However, human raters often rate more than one item, and humans are known to have diverging biases when rating data. Therefore, accounting for these dependencies on raters may lead to more accurate estimators. In future work, we will focus on modeling the behavior of individual raters to investigate the impact of rater disagreement on the amount of data needed for evaluation.
Specifically, we will use non-parametric bootstrapping and parametric methods such as those described in Figure \ref{fig:graphical} to study the impact of raters on evaluation.

We also aim to estimate response variance for multi-turn conversations and collect new data to externally validate the VET estimator's predictions for optimal choices of $N$ and $K$.

\section{Human-Centered AI Alignment}
\label{sec:alignment}

\subsection{Related Work}

\paragraph{Rater Disagreement} \citet{weerasooriya-etal-2023-vicarious} introduced \emph{vicarious offense} to tease apart disagreement in political discourse by asking raters how they think others would annotate the data. Raters with specific political leanings are asked whether they find the text offensive and whether they think people with other political leanings may find the text offensive. Such vicarious annotations can reveal whether groups can be trusted to reflect the opinions of other groups and affect rater recruitment, since fewer raters may be recruited from other groups.

CrowdTruth~\cite{Dumitrache_SAD_CROWDBIAS_HCOMP2018} introduced a set of metrics for capturing and interpreting rater disagreement in crowdsourcing, and the GRASP framework~\cite{prabhakaran-etal-2024-grasp} introduced metrics to understand the extent to which rater disagreement is based on group membership.

\paragraph{Learning Using Textual Feedback} 
Recent research has pivoted towards using LLMs themselves as scalable proxies for human judgment, serving as evaluators, critics, and sources of feedback~\cite{NEURIPS2023_zheng, pryzant-etal-2023-automatic, saunders_self-critiquing_2022}. This has given rise to sophisticated agentic frameworks that can detect errors, critique outputs, and iteratively refine them, particularly for tasks demanding factual correctness~\cite{akyurek-etal-2023-rl4f, NEURIPS2023_91edff07}. Methods like TextGrad~\cite{yuksekgonul2024textgrad} have even demonstrated how textual feedback can ``differentiate'' through complex systems to optimize performance.

\paragraph{Test-Time Scaling and Alignment}

Test-time scaling improves LLMs without weight modification and aims to enhance the performance of models by utilizing additional test-time compute resources~\cite{snell_scaling_2024, muennighoff_s1_2025, wang_sampling-efficient_2025}. These methods have also been applied to reward modeling, reinforcement learning, and alignment~\cite{hao2025rlthoughtsnavigatingllm, song2025rewardenoughllmsincontext, zhang2025unlockingrecursivethinkingllms}.

\subsection{Preliminary Work}

\myrq{4}{Are some demographic groups more cohesive than others when disclosing their own perceptions of offense and vicarious offense?}

We study the potential influence of political affiliation and demographics on raters' perception of offense~\cite{pandita-etal-2024-rater} to demonstrate the effect of group membership.
To this end, we use vicarious annotations~\cite{weerasooriya-etal-2023-vicarious} along with the GRASP framework~\cite{prabhakaran-etal-2024-grasp} and CrowdTruth~\cite{Dumitrache_SAD_CROWDBIAS_HCOMP2018}.

\paragraph{Experiments} We conduct our experiments on the VOICED dataset~\cite{weerasooriya-etal-2023-vicarious} containing YouTube comments labeled by diverse raters for personal and vicarious offense. We also use the toxicity ratings dataset~\citep{kumar2021designing}, comprising comments from Twitter labeled for toxicity by multiple raters. We consider political leaning and gender as dimensions to compare cohesion among different subgroups.

\paragraph{Results} The results (Tables \ref{tab:tb_ct_metrics}-\ref{tab:tb_vic_ct_metrics}) show that, of the political groups, Independents are the most cohesive, both with themselves and with others. Democrats are the least cohesive with others. Republicans are the least internally cohesive.

Our investigation into the dynamics of rater cohesion in politically charged content moderation settings, through the lens of self and \textit{vicarious annotation}, gender, and political affiliations, reveals valuable insights into the challenges of building inclusive and human-centered AI systems. Our findings reveal notable disparities in cohesion levels, highlighting the influence of gender and political affiliation.

\myrq{5}{Can textual feedback enhance the performance of LLMs during inference?}

We introduce \textbf{ProRefine} (Inference-time \textbf{Pro}mpt \textbf{Refine}ment with Textual Feedback)~\cite{pandita2026prorefine},
which focuses on optimizing the \textit{prompt}, a key element in chain-of-thought (CoT)~\citep{NEURIPS2022_wei} based LLM reasoning. ProRefine dynamically refines prompts for multi-step reasoning tasks without additional training or ground-truth labels.

\begin{figure}[ht]
    \centering
    \includegraphics[width=\linewidth]{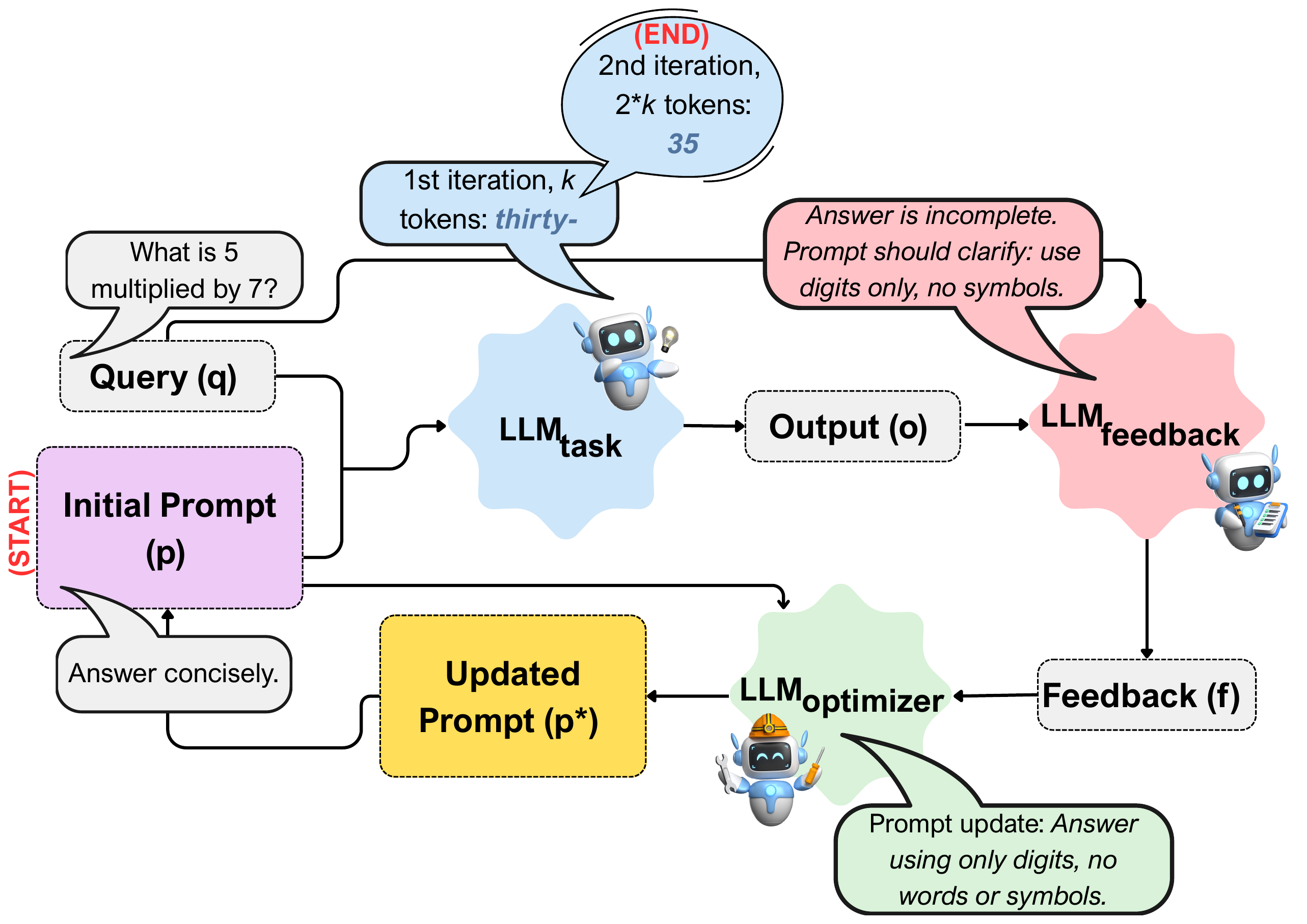}
    \caption{Overview of the ProRefine system, illustrating the iterative process of prompt optimization using feedback from LLMs. In each iteration, $LLM_{task}$ extends its output by an additional $k$ tokens, enabling step-by-step feedback to progressively refine the prompt with $LLM_{optimizer}$.}
    \label{fig:system_diagram}
\end{figure}

\paragraph{Methods} ProRefine adaptively improves prompts for a task-performing LLM ($LLM_{task}$) by using textual feedback from a judge ($LLM_{feedback}$) and an optimizer ($LLM_{optimizer}$). This workflow (Figure \ref{fig:system_diagram}), motivated by the teacher-student framework~\cite{torrey2013teaching} where a teacher agent guides a student agent to perform a task by providing feedback at intermediate steps, but implemented via LLM interactions without pre-training, represents a novel approach to adaptive agentic reasoning. ProRefine involves interactions between three LLMs:

\noindent\bm{$LLM_{task}$}: Executes the task based on the current prompt, generating the initial and subsequent outputs.

\noindent\bm{$LLM_{feedback}$}: A model that critiques the $LLM_{task}$'s output, providing detailed feedback on improvements. This model should be capable of providing insightful and accurate critiques~\cite{bai_constitutional_2022, saunders_self-critiquing_2022}.

\noindent\bm{$LLM_{optimizer}$}: Interprets the feedback and refines the prompt, aiming for coherent and task-focused improvements. This LLM is crucial for ensuring the prompt evolves effectively.

\paragraph{ProRefine} (Algorithm \ref{alg:prorefine}) works as follows:

\paragraph{Initialization:}

Start with an initial prompt $p$ for the task, a query $q$, and parameters defining the generation and optimization process ($k$ tokens per step, $n$ maximum steps).

\paragraph{Generation and Feedback Loop:}
\begin{itemize}
    \item \textbf{Generation:} Use $LLM_{task}$ to generate an output based on the current prompt $p^*$ and query $q$. This step is limited to $i$ $*$$k$ tokens to control the granularity of the feedback. In each iteration, $LLM_{task}$ produces $k$ more tokens, attempting to refine prior output while progressively continuing its response to the query.
    \item \textbf{Feedback:} $LLM_{feedback}$ evaluates the generated output $o_i$ against the query $q$ to provide textual feedback $f_i$. This feedback encapsulates how the output could be improved, focusing on aspects such as accuracy, relevance, or coherence.
    \item \textbf{Optimization:} $LLM_{optimizer}$ uses the feedback $f_i$ to refine the prompt $p^*$. This step involves modifying the prompt to better align with the task requirements or to correct identified deficiencies in previous generations.
\end{itemize}

\paragraph{Termination:}
The process iterates until either the maximum number of steps $n$ is reached or an end-of-sequence (EOS) token is detected in the output, indicating the completion of the task.

\paragraph{Experiments} We conduct the experiments on five reasoning tasks -- object counting and word sorting from the BIG-Bench Hard benchmark~\cite{srivastava2023beyond}, grade-school math problem-solving from GSM8K~\cite{cobbe2021training}, math word problems from SVAMP~\cite{patel-etal-2021-nlp}, and algebraic word problems from AQUARAT~\cite{ling-etal-2017-program}, each of which involves multi-step reasoning.

We experiment with three models - \textit{Llama3.2-1B-instruct}, \textit{Llama3.2-3B-instruct}, and \textit{Llama3.1-8B-instruct}~\cite{llama3_2_model_card} for $LLM_{task}$. The prompts are optimized using Algorithm \ref{alg:prorefine}, with \textit{Llama3.1-70B-instruct} used for feedback generation, prompt optimization, and evaluation. We select the values of hyperparameters $k=10$ and $n=25$ to control the granularity of feedback and duration of optimization. Hyperparameters $k$ and $n$ were fixed based on general preliminary exploration and not tuned per task using benchmark training/validation data. We compare ProRefine against the zero-shot Chain-of-Thought (CoT) baseline and TextGrad~\cite{yuksekgonul2024textgrad}, and report test accuracy with a 95\% confidence interval.

\paragraph{Results} Our results (Table \ref{tab:results}) demonstrate that ProRefine significantly improves $LLM_{task}$ performance over the zero-shot CoT baseline in all but one experiment, and it outperforms TextGrad in 11 out of 15 cases overall. For \textit{Llama3.2-1B-instruct} model, ProRefine can significantly outperform CoT and TextGrad on 2 out of 5 datasets. For \textit{Llama3.2-3B-instruct} model, ProRefine can outperform CoT and TextGrad on 3 out of 5 datasets with one significant result. For \textit{Llama3.1-8B-instruct} model, ProRefine can outperform CoT and TextGrad on all 5 datasets with 4 significant results.

\subsection{Proposed Work}

\myrq{6}{How can we effectively adapt LLMs to represent pluralistic perspectives?}

In the future, we aim to develop methods to align model outputs across different perspectives by leveraging textual feedback from multiple feedback models that represent specific perspectives. Such feedback models may be prompted to adopt a different persona or fine-tuned to provide feedback from a specific perspective. Then, we can use the feedback to either optimize the prompt for specific perspectives or have the task-performing LLM adapt its output directly. In the first case, perspective-optimized prompts can be used to generate outputs from different perspectives, which are then synthesized to produce a single output. These approaches can refine LLM outputs at inference time, making them suitable for black-box scenarios where model weights are inaccessible and for agentic workflows. In the second case, the task-performing LLM may also be fine-tuned to utilize feedback from different perspectives.

In addition to relying on textual feedback from perspectivist models, we also aim to investigate whether textual or numeric feedback is more effective for alignment. To this end, we will train multiple reward models for different perspectives and leverage them in a conventional RLHF setting.

\paragraph{Bandits for Prompt Optimization} For prompt optimization, we aim to leverage bandit algorithms to select the best-performing prompts for a given task, ideally under a budget constraint. First, $m$ prompt candidates may be generated for a given task, and then each of these prompts will be used to generate $n$ outputs from a task-performing LLM. The generated outputs will each receive feedback (textual or numeric) based on the output quality, and the prompts may be refined based on this feedback. In the subsequent round, we may have $m\times n$ candidate prompts, and the process continues iteratively. Bandit algorithms are well-suited for this scenario to keep track of the better-performing prompts and eventually pick the best-performing prompt under a given budget.

\section{Conclusion}
\label{sec:conclusion}

This PhD thesis proposal highlights the need for integrating diverse perspectives for reproducible machine learning evaluation and pluralistic alignment. For reproducible ML evaluations, we experimented with simulations based on real-world datasets and showed that these datasets lack enough responses per item. We also investigated the critical trade-off between the number of items and the number of responses per item for achieving reliable machine learning model evaluation under a fixed budget. Our findings demonstrate that increasing the number of responses per item is often a more effective strategy for achieving reliable evaluation than increasing the number of items. Furthermore, we established that this trade-off is dependent on the metric. Methods developed as a part of this proposal, such as ProRefine, have demonstrated significant performance improvements on multi-step reasoning tasks. Future research aims to leverage these findings to align model outputs across different perspectives by utilizing LLM-generated textual feedback.

Our research provides a clear, data-driven methodology for ML practitioners to design more effective and budget-conscious evaluations. By moving towards a perspective-aware paradigm and strategically collecting multiple responses, the field can build greater trust and confidence in model performance. Our findings regarding the dynamics of rater cohesion further underscore the importance of building human-centered AI systems and open up a strategic avenue for more efficient rater recruitment.

\section*{Limitations}

The effectiveness of the VET simulator depends on how well the probabilistic models capture realistic distributions of responses over items. Although we used rigorous methods to fit the parameters of these distributions to our datasets, our choice of distribution family to use for each dataset was based on visual inspection of the data in the case of regression tasks. One key limitation future work will address is that we treat the responses as independent from item-to-item, when in reality responses usually depend on which human annotator or instance of a model produced the response. Hypothesis testing such as that described here is not a comprehensive measure of data quality; it only estimates the likelihood of sampling error. It does not account for sampling bias, leading to data that is not representative of the sampling distribution.

Our findings about rater cohesion may not be generalizable to other demographics such as education level, cultural background, and economic status. Future studies should employ the proposed framework to investigate the level of cohesion among raters belonging to other important demographic subgroups. Another limitation of this work is the simplification of political ideologies into three groups: Democrats, Republicans, and Independents. This, however, may not capture the full spectrum of political beliefs and identities. For instance, a rater can be socially Republican but fiscally Liberal. A more granular analysis that considers the multidimensional nature of political ideologies could reveal more intricate patterns of cohesion.

While ProRefine is designed for cost-effective hybrid deployments, its iterative process inherently increases inference-time latency and computational cost compared to a single-pass query. Our evaluation is currently focused on mathematical and multi-step reasoning tasks. Further research is needed to assess performance across a broader range of reasoning tasks and domains. The iterative nature of ProRefine lacks a formal convergence guarantee. In some cases, the refinement process can suffer from prompt degradation after many iterations or plateau before reaching an optimal solution. Investigating methods to ensure stable and monotonic improvement is a key area for future research.


\bibliography{anthology-1,anthology-2,references,custom}

\appendix

\section{Appendix}
\label{sec:appendix}

\subsection{Reproducible ML Evaluation}

\begin{figure}[h]
\centering
\begin{subfigure}[]{\linewidth}
\centering
\includegraphics[width=\linewidth]{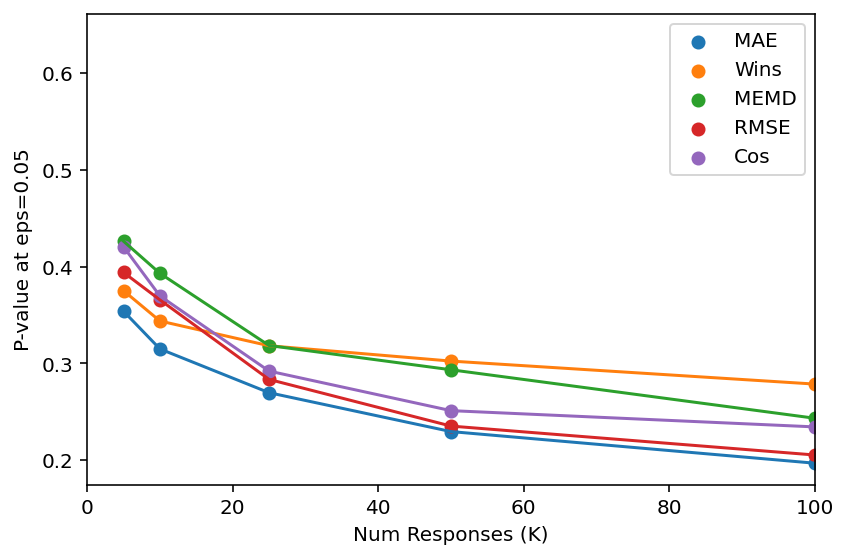}
\caption{Toxicity ($\epsilon=0.05$)}
\label{fig:toxicity_various_metrics}
\end{subfigure}
\hspace{.3cm}
\begin{subfigure}[]{\linewidth}
\centering
\includegraphics[width=\linewidth]{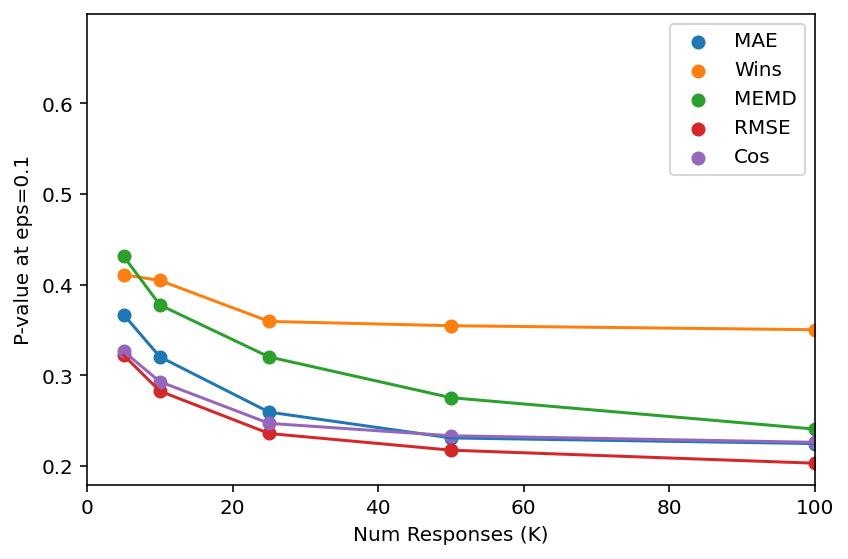}
\caption{MultiDomain ($\epsilon=0.1$)}
\label{fig:mdagreement_various_metrics}
\end{subfigure}
\caption{
\pv\ vs $K$ with a fixed budget $N \times K = 2500$ for various metrics. Each data point is estimated from $10,000$ samples.
}
\label{fig:various_metrics}
\end{figure}

\subsubsection{Power Analysis}
\label{sec:power_analysis}

\begin{figure}[h]
\centering
\begin{subfigure}[]{\linewidth}
\centering
\includegraphics[width=\linewidth]{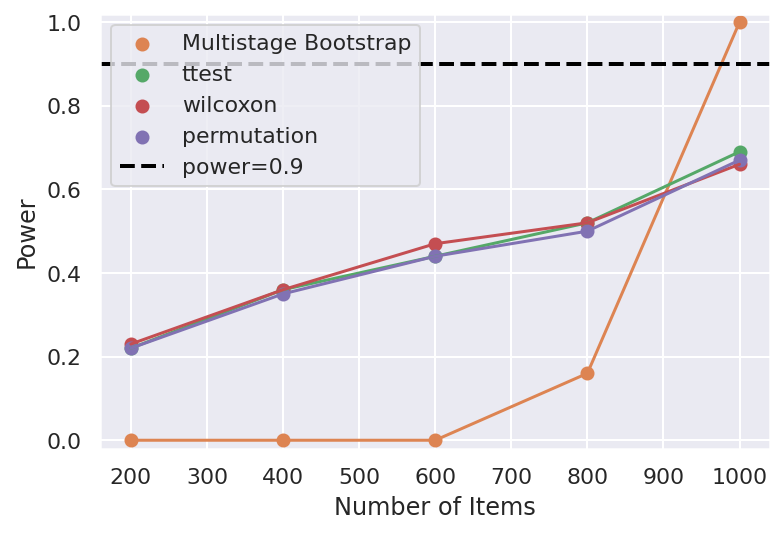}
\caption{Varying $N$ with $K=5$}
\end{subfigure}
\begin{subfigure}[]{\linewidth}
\centering
\includegraphics[width=\linewidth]{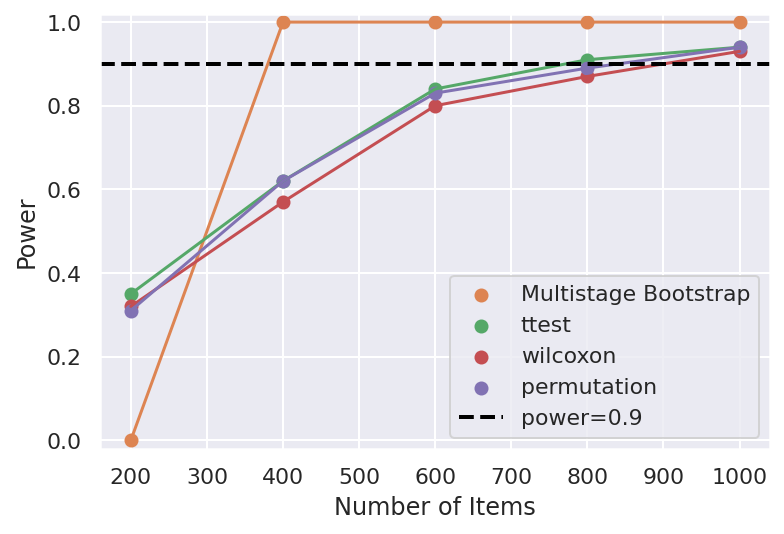}
\caption{Varying $N$ with $K=10$}
\end{subfigure}
\begin{subfigure}[]{\linewidth}
\centering
\includegraphics[width=\linewidth]{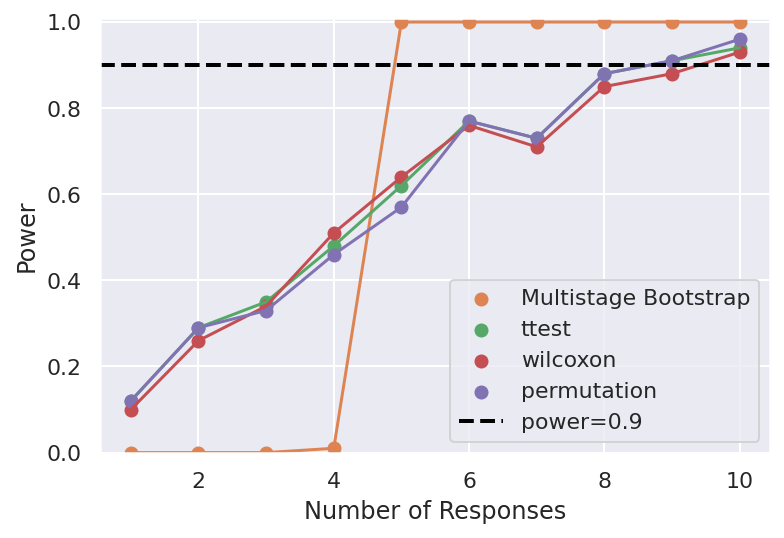}
\caption{Varying $K$ with $N=1000$}
\end{subfigure}
\caption{Power Analysis of Toxicity data ($\epsilon=0.1)$. Each data point is estimated from 1000 outer-level samples, each consisting of 10000 inner-level samples.}
\label{fig:toxicity_power_vs_sample_size}
\end{figure}

Figure~\ref{fig:toxicity_power_vs_sample_size}
demonstrate greater statistical power for Multistage Bootstrap as sample size
with respect to either number of items or responses
increases, achieving a power of 90\% (i.e., probability of not rejecting the null hypothesis when it's false) before baseline hypothesis tests.
As usual, we use $\alpha=0.05$ as the significance level for power calculation, i.e., the data is inconsistent with the null hypothesis at least 95\% of the time.
While the power of all these tests benefit from having more responses, the rate of improvement is markedly more rapid for Multistage Bootstrap.

\subsubsection{Algorithms for Categorical Simulations}

\begin{algorithm}
\SetKwInput{KwInput}{Input parameters} 
\caption{Simulations for $H_{alt}$}
\label{alg:alt}
\KwInput{$N, K, M, \alpha, \rho, \epsilon$}
\For{{$i=1$ \KwTo $N$}}{
\tcp{sample categorical parameters}
$\bm{\beta_i} = \beta_{i,1},...,\beta_{i,M} \sim Dir(\alpha_{1},...,\alpha_{M})$ \;
\tcp{sample noise parameters}
$\bm{\varrho_i} = \varrho_{i,1},...,\varrho_{i,M} \sim Dir(\rho_{1},...,\rho_{M})$ \;
\tcp{convex combination of categorical \& noise parameters}
$\bm{\gamma_i} = (1-\epsilon)*\bm{\beta_i} + \epsilon*\bm{\varrho_i}$ \;
\tcc{sample $j$’s response to $i$}
    \tcp{Gold}
    \For{{$j=1$ \KwTo $k$}}{
        $G_{i,j} = Cat(\beta_{i,1},...,\beta_{i,M})$ \;
    }
    \tcp{Model A}
    \For{{$j=1$ \KwTo $K$}}{
        $A_{i,j} = Cat(\beta_{i,1},...,\beta_{i,M})$ \;
    }
    \tcp{Model B}
    \For{{$j=1$ \KwTo $K$}}{
        $B_{i,j} = Cat(\gamma_{i,1},...,\gamma_{i,M})$ \;
    }
}
\end{algorithm}

\begin{algorithm}
\SetKwInput{KwInput}{Input parameters} 
\caption{Simulations for $H_{null}$}
\label{alg:null}
\KwInput{$N, K, M, \alpha, \rho, \epsilon$}
\For{{$i=1$ \KwTo $N$}}{
\tcp{Use same steps as Algorithm \ref{alg:alt} for $\bm{\beta_i}$, $\bm{\varrho_i}$, $\bm{\gamma_i}$ and $G_{i,j}$}
    \tcp{Model A}
    \For{{$j=1$ \KwTo $K$}}{
        $x \sim Bernoulli(0.5)$ \;
        \If{$x==0$}{
            $A_{i,j} = Cat(\beta_{i,1},...,\beta_{i,M})$ \;
        }
        \Else{
            $A_{i,j} = Cat(\gamma_{i,1},...,\gamma_{i,M})$ \;
        }
    }
    \tcp{Model B}
    \For{{$j=1$ \KwTo $K$}}{
        $x \sim Bernoulli(0.5)$ \;
        \If{$x==0$}{
            $B_{i,j} = Cat(\gamma_{i,1},...,\gamma_{i,M})$ \;
        }
        \Else{
            $B_{i,j} = Cat(\beta_{i,1},...,\beta_{i,M})$ \;
        }
    }
}
\end{algorithm}

\begin{algorithm}
\SetKwInput{KwInput}{Input}
\caption{Calculate \pv\ }
\label{alg:pval}
\KwInput{$\Gamma^{alt}, \Gamma^{null}$}
$p \leftarrow 0$ \;
\For{{$score \in \Gamma^{alt}$}}{
    $p \leftarrow p + (|\Gamma^{null}>score|)/(|\Gamma^{null}|)$ \;
}
$p \leftarrow p/|\Gamma^{alt}|$ \;
\end{algorithm}

\begin{algorithm}
\SetKwInput{KwInput}{Input}
\caption{Calculate Confidence Interval (CI)}
\label{alg:ci}
\KwInput{$\Gamma^{alt}$}
$\hat{\Gamma} \leftarrow mean(\Gamma^{alt})$\;
$\Gamma^{alt}_{sorted} \leftarrow \text{sort}(\Gamma^{alt})$\;
\tcp{Choose 2.5th and 97.5th percentile (95\% CI)}
$\text{CI}_{lower} \leftarrow 2\hat{\Gamma} - \Gamma^{alt}_{sorted}[975]$\;
$\text{CI}_{upper} \leftarrow 2\hat{\Gamma} - \Gamma^{alt}_{sorted}[25]$\;
$\text{CI} \leftarrow [\text{CI}_{lower}, \text{CI}_{upper}]$\;
\end{algorithm}

\subsection{Rater Cohesion}

\begin{table*}
\centering
\small
\begin{tabular}{r|ccccccc}
& & & & Cross $\otimes$ & Plurality $\cap$ & Voting $\otimes$ & \\
Group & IRR $\cap$ & XRR $\otimes$ & Negentropy $\cap$ & Negentropy & Size & Agreement & GAI \\
\hline
Dem & $\uparrow$0.238 & $\downarrow$0.197 & $\downarrow$0.403 & $\downarrow$0.349 \cellcolor{orange} & $\downarrow$0.855 & \textbf{$\downarrow$0.367} \cellcolor{green} & \textbf{$\uparrow$1.203} \cellcolor{cyan} \\
Rep & $\downarrow$0.167 & $\downarrow$0.193 & $\downarrow$0.376 & $\uparrow$0.381 & $\downarrow$0.851 & $\downarrow$0.473 & $\downarrow$0.864 \\
Ind & $\uparrow$0.251 & $\uparrow$0.215 & \textbf{$\uparrow$0.487} \cellcolor{green} & $\uparrow$0.383 & \textbf{$\uparrow$0.898} \cellcolor{green} & $\uparrow$0.537 \cellcolor{orange} & $\uparrow$1.165 \\
\midrule
Men & $\uparrow$0.213 & \textbf{$\downarrow$0.187} \cellcolor{green} & $\uparrow$0.387 & $\downarrow$0.384 & $\uparrow$0.861 & $\downarrow$0.493 & $\uparrow$1.141 \\
Women & $\downarrow$0.202 & \textbf{$\downarrow$0.187} \cellcolor{green} & $\downarrow$0.379 & $\uparrow$0.384 & $\downarrow$0.854 & $\downarrow$0.482 & $\uparrow$1.085 \\
\hline
Dem, Men & $\uparrow$0.204 & $\uparrow$0.205 & $\downarrow$0.484 & $\uparrow$0.359 & $\downarrow$0.884 & $\downarrow$0.340 & $\uparrow$0.993 \\
Dem, Women & \textbf{$\uparrow$0.305} \cellcolor{cyan} & $\uparrow$0.222 & $\downarrow$0.507 & $\downarrow$0.302 \cellcolor{orange} & $\downarrow$0.892 \cellcolor{orange} & \textbf{$\downarrow$0.206} \cellcolor{green} & \textbf{$\uparrow$1.373} \cellcolor{cyan} \\
Rep, Men & $\downarrow$0.148 & $\downarrow$0.197 & $\uparrow$0.481 & $\uparrow$0.371 & $\uparrow$0.885 & $\uparrow$0.371 & $\downarrow$0.750 \\
Rep, Women & $\downarrow$0.175 & \textbf{$\downarrow$0.154} \cellcolor{green} & \textbf{$\downarrow$0.433} \cellcolor{cyan} & $\downarrow$0.299 & \textbf{$\downarrow$0.864} \cellcolor{cyan} & $\downarrow$0.272 & $\uparrow$1.142 \cellcolor{orange} \\
Ind, Men & $\uparrow$0.284 & $\uparrow$0.241 & $\uparrow$0.537 & $\downarrow$0.348 & $\uparrow$0.910 & $\uparrow$0.349 & $\uparrow$1.182 \\
Ind, Women & \textbf{$\downarrow$0.110} \cellcolor{cyan} & $\downarrow$0.174 & $\uparrow$0.572 \cellcolor{orange} & \textbf{$\uparrow$0.423} \cellcolor{green} & $\uparrow$0.930 \cellcolor{orange} & $\uparrow$0.393 & \textbf{$\downarrow$0.631} \cellcolor{cyan} \\
\hline
$\Delta$ & 0.047 & 0.041 & 0.053 & 0.083 & 0.029 & 0.060 & 0.130 \\
\end{tabular}
\caption{Results of in-group and cross-group cohesion metrics on $\mathcal{D}_\textit{voiced}$ after CrowdTruth (CT) filtering. $\cap$ stands for in-group metric and $\otimes$ stands for cross-group metric. Significant results are indicated in bold at the $p=0.05$ significance level, $\downarrow$ indicates the result is less than expected under the null hypothesis, and $\uparrow$ indicates the result is greater than expected. \colorbox{orange}{Orange} indicates the result is significant before applying CT, \colorbox{cyan}{Cyan} indicates the result is significant after applying CT, and \colorbox{green}{Green} indicates the result is significant before and after applying CT. $\Delta$ is the mean absolute difference of metric scores before and after applying CT.}
\label{tab:tb_ct_metrics}
\end{table*}

\begin{table*}
\centering
\small
\begin{tabular}{r|ccccccc}
& & & & Cross $\otimes$ & Plurality $\cap$ & Voting $\otimes$ & \\
Group & IRR $\cap$ & XRR $\otimes$ & Negentropy $\cap$ & Negentropy & Size & Agreement & GAI \\
\hline
Rep $\rightarrow$ Dem (v Dem) & $\downarrow$0.181 & $\downarrow$0.176 & \textbf{$\downarrow$0.419} \cellcolor{green} & $\downarrow$0.411 & \textbf{$\downarrow$0.871} \cellcolor{green} & $\downarrow$0.331 & $\downarrow$1.027 \\
Ind $\rightarrow$ Dem (v Dem) & $\uparrow$0.252 & \textbf{$\uparrow$0.231} \cellcolor{cyan} & $\downarrow$0.502 & $\downarrow$0.423 & $\uparrow$0.906 & \textbf{$\uparrow$0.418} \cellcolor{cyan} & $\downarrow$1.091 \\
\hline
Dem $\rightarrow$ Rep (v Rep) & $\uparrow$0.230 & $\downarrow$0.166 & \textbf{$\downarrow$0.376} \cellcolor{green} & \textbf{$\downarrow$0.346} \cellcolor{green} & \textbf{$\downarrow$0.840} \cellcolor{green} & $\uparrow$0.283 & $\uparrow$1.389 \\
Ind $\rightarrow$ Rep (v Rep) & $\uparrow$0.215 & $\uparrow$0.191 & $\uparrow$0.470 & $\downarrow$0.402 & $\uparrow$0.887 & \textbf{$\uparrow$0.393} \cellcolor{green} & $\downarrow$1.123 \\
\hline
Dem $\rightarrow$ Ind (v Ind) & $\uparrow$0.203 & $\uparrow$0.200 & $\downarrow$0.413 \cellcolor{orange} & \textbf{$\uparrow$0.487} \cellcolor{green} & $\downarrow$0.860 \cellcolor{orange} & $\uparrow$0.353 & $\uparrow$1.016 \\
Rep $\rightarrow$ Ind (v Ind) & $\downarrow$0.164 & $\uparrow$0.200 & \textbf{$\downarrow$0.393} \cellcolor{cyan} & \textbf{$\uparrow$0.486} \cellcolor{green} & $\downarrow$0.857 & $\uparrow$0.372 & $\downarrow$0.821 \\
\hline
$\Delta$ & 0.036 & 0.039 & 0.055 & 0.073 & 0.029 & 0.046 & 0.060 \\
\end{tabular}
\caption{Results of vicarious alignment on $\mathcal{D}_\textit{voiced}$ after CrowdTruth (CT) filtering. $\cap$ stands for in-group metric and $\otimes$ stands for cross-group metric. Significant results are indicated in bold at the $p=0.05$ significance level, $\downarrow$ indicates the result is less than expected under the null hypothesis, and $\uparrow$ indicates the result is greater than expected. \colorbox{orange}{Orange} indicates the result is significant before applying CT, \colorbox{cyan}{Cyan} indicates the result is significant after applying CT, and \colorbox{green}{Green} indicates the result is significant before and after applying CT. $\Delta$ is the mean absolute difference of metric scores before and after applying CT.}
\label{tab:tb_vic_ct_metrics}
\end{table*}

\subsection{ProRefine}

\begin{algorithm}
\SetKwInput{KwInput}{Input} 
\SetKwInput{KwOutput}{Output} 
\SetKw{Break}{break}
\caption{ProRefine}
\label{alg:prorefine}
\KwInput{Query: $q$, Initial prompt: $p$, tokens\_per\_step: $k$, max\_steps: $n$, LLMs: $LLM_{task}$, $LLM_{feedback}$, $LLM_{optimizer}$}
\KwOutput{Optimized prompt: $p^*$}
$p^*=p$\\
\For{{$i=1$ \KwTo $n$}}{
$o_i = LLM_{task}(p^*, q)$ \tcp{Generate $i*k$ tokens}
$f_i = LLM_{feedback}(q, o_i)$ \tcp{Get textual feedback}
$p* = LLM_{optimizer}(p^*, f_i)$ \tcp{Optimize the prompt}
\If{$EOS\_token$ in $o_i$}{\Break}
}
\Return $p^*$ \tcp{Return final optimized prompt}
\end{algorithm}

\begin{table*}
\centering
\small
\begin{tabular}{l|l|c|c|c}
\textbf{Dataset} & \textbf{Method} & \textbf{Llama-3.2 1B-it} & \textbf{Llama-3.2 3B-it} & \textbf{Llama-3.1 8B-it}\\
\hline

\multirow{3}{*}{Object}
& CoT                 & 0.48 [0.382, 0.578] & 0.65 [0.556, 0.744] & 0.73 [0.643, 0.817] \\
& TextGrad           & \textbf{0.62} [0.524, 0.716] & 0.73 [0.643, 0.817] & 0.86 [0.792, 0.928] \\
Counting & ProRefine (no verifier)       & 0.51 [0.412, 0.608] & \textbf{0.75} [0.665, 0.835] & 0.77 [0.687, 0.853] \\
& \cellcolor{gray!20}ProRefine (verifier)    & \cellcolor{gray!20}0.6 [0.503, 0.696] & \cellcolor{gray!20}0.72 [0.632, 0.808] & \cellcolor{gray!20}\textbf{0.89}* [0.839, 0.959] \\
\cline{2-5}
& \textsuperscript{\textdagger}ProRefine (optimal verifier)    & 0.67 [0.577, 0.763] & 0.85* [0.780, 0.920] & 0.94* [0.893, 0.987] \\
\hline

\multirow{3}{*}{Word}
& CoT               & 0.11 [0.048, 0.172] & 0.10 [0.041, 0.159] & 0.50 [0.401, 0.598] \\
& TextGrad         & \textbf{0.33}* [0.237, 0.423] & \textbf{0.61}* [0.514, 0.706] & 0.69* [0.599, 0.781] \\
Sorting & ProRefine (no verifier)     & 0.22 [0.138, 0.302] & 0.47* [0.372, 0.568] & 0.68 [0.595, 0.779] \\
& \cellcolor{gray!20}ProRefine (verifier)  & \cellcolor{gray!20}0.19 [0.113, 0.267] & \cellcolor{gray!20}0.32* [0.228, 0.412] & \cellcolor{gray!20}\textbf{0.71}* [0.621, 0.799] \\
\cline{2-5}
& \textsuperscript{\textdagger}ProRefine (optimal verifier)  & 0.29* [0.192, 0.368] & 0.53* [0.432, 0.628] & 0.86** [0.792, 0.928] \\
\hline

\multirow{5}{*}{GSM8K}
& CoT               & 0.450 [0.423, 0.476] & 0.809 [0.787, 0.829] & 0.819 [0.797, 0.839] \\
& TextGrad         & 0.463 [0.436, 0.489] & 0.801 [0.779, 0.822] & 0.864* [0.845, 0.882] \\
& ProRefine (no verifier)     & 0.636** [0.610, 0.662] & 0.797 [0.774, 0.818] & 0.843 [0.823, 0.863 \\
& \cellcolor{gray!20}ProRefine (verifier)  & \cellcolor{gray!20}\textbf{0.654}** [0.627, 0.678] & \cellcolor{gray!20}\textbf{0.866}** [0.847, 0.883] & \cellcolor{gray!20}\textbf{0.885}* [0.868, 0.902] \\
\cline{2-5}
& \textsuperscript{\textdagger}ProRefine (optimal verifier)  & 0.725** [0.701, 0.749] & 0.904** [0.888, 0.920] & 0.936** [0.922, 0.949] \\
\hline

\multirow{5}{*}{SVAMP}
& CoT               & 0.689 [0.66, 0.718] & 0.869 [0.848, 0.890] & 0.854 [0.832 , 0.876] \\
& TextGrad         & 0.684 [0.655, 0.713] & 0.861 [0.840, 0.882] & 0.84 [0.817, 0.863] \\
& ProRefine (no verifier)     & 0.774** [0.748, 0.800] & 0.878 [0.858, 0.898] & 0.877 [0.857, 0.897] \\
& \cellcolor{gray!20}ProRefine (verifier)  & \cellcolor{gray!20}\textbf{0.808}** [0.784, 0.832] & \cellcolor{gray!20}\textbf{0.896} [0.877, 0.915] & \cellcolor{gray!20}\textbf{0.893}* [0.874, 0.912] \\
\cline{2-5}
& \textsuperscript{\textdagger}ProRefine (optimal verifier)  & 0.861** [0.840, 0.882] & 0.925** [0.909, 0.941] & 0.938** [0.923, 0.953] \\
\hline

\multirow{5}{*}{AQUARAT}
& CoT               & 0.259 [0.202, 0.31] & \textbf{0.563} [0.498, 0.620] & 0.586 [0.522, 0.643] \\
& TextGrad         & \textbf{0.311} [0.250, 0.364] & 0.524 [0.462 , 0.585] & 0.559 [0.494, 0.616] \\
& ProRefine (no verifier)     & 0.205 [0.151, 0.250] & 0.343 [0.284, 0.401] & 0.398 [0.337 , 0.458] \\
& \cellcolor{gray!20}ProRefine (verifier)  & \cellcolor{gray!20}0.268 [0.209, 0.318] & \cellcolor{gray!20}0.551 [0.486 , 0.608] & \cellcolor{gray!20}\textbf{0.606} [0.542, 0.663] \\
\cline{2-5}
& \textsuperscript{\textdagger}ProRefine (optimal verifier)  & 0.354 [0.292, 0.409] & 0.598 [0.538, 0.659] & 0.657 [0.595, 0.712 ] \\

\end{tabular}
\caption{Test Accuracy with 95\% confidence intervals across five benchmark datasets and models. * and ** denote statistically significant improvements over one or two baseline methods, respectively. Results in bold indicate the highest accuracy for a dataset-method combination. \textsuperscript{\textdagger} demonstrates the upper bound potential of the optimization loop and the impact of verifier quality. \textit{Llama3.1-70B-instruct} is employed for feedback generation, prompt optimization, and evaluation.}
\label{tab:results}
\end{table*}

\end{document}